\documentclass{article}
\usepackage{nips07submit_e,times}
\usepackage{epsfig}
\usepackage{graphicx}
\usepackage{amsmath,amsfonts,verbatim}
\usepackage{amssymb}
\usepackage[pagebackref=true,breaklinks=true,letterpaper=true,colorlinks,bookmarks=false]{hyperref}

\title{Bounding the Probability of Error for High Precision Recognition}

\author{
Andrew Kae\\
Department of Computer Science\\
University of Massachusetts Amherst\\
Amherst, MA 01003\\
\texttt{akae@cs.umass.edu} \\
\And
Gary B. Huang\\
Department of Computer Science\\
University of Massachusetts Amherst\\
Amherst, MA 01003\\
\texttt{gbhuang@cs.umass.edu} \\
\And
Erik Learned-Miller\\
Department of Computer Science\\
University of Massachusetts Amherst\\
Amherst, MA 01003\\
\texttt{elm@cs.umass.edu} \\
\\
Technical Report UM-CS-2009-031\\
June 30, 2009
}

\begin{document}

\maketitle

\begin{abstract}
We consider models for which it is important, early in processing, to
estimate some variables with high precision, but perhaps at
relatively low rates of recall. If some variables can be identified
with near certainty, then they can be conditioned upon, allowing
further inference to be done efficiently.  Specifically, we consider
optical character recognition (OCR) systems that can be bootstrapped
by identifying a subset of correctly translated document words with very high precision. This ``clean
set'' is subsequently used as document-specific training
data.  While many current OCR systems produce measures of confidence for
the identity of each letter or word, thresholding these confidence
values, even at very high values, still produces some errors.

We introduce a novel technique for identifying a set of correct words
with very high precision. Rather than estimating posterior
probabilities, we {\bf bound} the probability that any given word is
incorrect under very general assumptions, using an approximate worst
case analysis.  As a result, the parameters of the model are nearly
irrelevant, and we are able to identify a subset of words, even in noisy
documents, of which we are highly confident. On our set of 10
documents, we are able to identify about 6\% of the words on average
{\em without making a single error.} This ability to produce word
lists with very high precision allows us to use a family of
models which depends upon such clean word lists. 


\end{abstract}

\section{Introduction}
The goal of this paper is to post-process the output from an optical
character recognition (OCR) program and identify a list of words which
the program got correct. The motivation is to use these correct words
to build new OCR models based upon training data from the document
itself, i.e., to build a document-specific model.

In this paper, we focus on the first step--identifying the 
correct words in someone else's OCR program output. While this may seem
like an easy thing to do, to our knowledge, there are no existing
techniques to perform this task with very high accuracy. There are
many methods that could be used to produce lists of words that
are mostly correct, but contain some errors. Unfortunately, such
lists are not much good as training data for document-specific models
since they contain errors, and these errors in training propagate
to create more errors later.

Thus, it is essential that our error rate be very low in the list of
words we choose as correct. As described below, we do not in fact
make any errors at all in our generated lists, which makes them 
appropriate for training document specific models. Before going into
the details of our method and how we achieve an empirical error
rate of 0\% for this task, we give some background on why we believe
this problem and our approach are interesting. 

\section{Background}
Humans and machines both make lots of errors in recognition problems.
However, one of the most interesting differences between people and
machines is that, for some inputs, humans are extremely confident of
their results and appear to be well-justified in this confidence.
Machines, on the other hand, while producing numbers such as posterior
probabilities which are supposed to represent confidences, are often
wrong even when posterior probabilities are extremely close to 1.

This is a particularly vexing problem when using generative models in
areas like computer vision and pattern recognition. Consider a two
class problem in which we are discriminating between two similar image
classes, $A$ and $B$. Because images are so high-dimensional,
likelihood exponents are frequently very small, and small errors in
these exponents can render the posteriors meaningless. For example,
suppose that $Pr(\tt{image}|A)=\exp(-1000+\epsilon_A)$ and
$Pr(\tt{image}|B)=\exp(-1005+\epsilon_B)$, where $\epsilon_A$ and
$\epsilon_B$ represent errors in the estimates of the image
distributions.\footnote{Such errors are extremely difficult to avoid
  in high-dimensional estimation problems, since there is simply not
  enough data to estimate the exponents accurately.} Assuming an equal
prior on $A$ and $B$, if $\epsilon_A$ and $\epsilon_B$ are Gaussian
distributed with standard deviation similar to the differences in the
exponents, then we will frequently conclude, incorrectly, that
$Pr(B|\tt{image})\approx 1$ and $Pr(A|\tt{image})\approx 0$.
This phenomenon, which is quite common in computer vision, makes it
quite difficult to assess confidence values in recognition problems.

As an alternative to estimating posterior probabilities very
accurately in order to be sure of certain results, we suggest an
alternative. We formulate our confidence estimate as an hypothesis
test that a certain result is {\em incorrect}, and if there is
sufficient evidence, we reject the hypothesis that the result is
incorrect. As we shall see, this comes closer to {\em bounding} the
probabilities of certain results, which can be done with greater
confidence, than {\em estimating} the probability of results, which is
much more difficult. A critical aspect of our approach is that if
there is insufficient evidence to reject a hypothesis, then {\em we
  make no decision at all.} Our process only makes decisions when
there is enough evidence, and avoids making decisions when there is
not. More formally, we search for cases in which our conservative
bounds on the probability of error are non-vacuous--if they are
vacuous, we make no decision.

\subsection{OCR and Document Specific Modeling}
Despite claims to the contrary, getting OCR systems to obtain very
high accuracy rates on moderately degraded documents continues to be a
challenging problem \cite{Nagy2000}.  One promising approach to
achieving very high OCR accuracy rates is to incorporate {\em document specific
  modeling}
\cite{Ho98bootstrappingtext,HongHull1995,HongHull_SPIE_1995}. This set
of approaches attempts to refine OCR models to specifically model the
document currently being processed by adapting to the fonts in the
document, adapting to the noise model in the document, or adapting to
the lexicon in the document.

If one had some method for finding a sample of words in a document
which were known to be correct with high confidence, one could
effectively use the characters in such words as training data with
which to build document specific models of the fonts in a document.
This chicken-and-egg problem is not easy to solve, however.  {\bf Our
  main result in this paper is to present a method for post-processing
  third party OCR outputs, in moderately difficult documents, to
  produce lists of words that are 100\% correct.}
As a comparison with our results, we do experiments using a public
domain OCR system that maintains a built-in confidence measure for its
own outputs. We show that there is no threshold on this confidence
value that produces clean word lists for most documents without either
an unacceptably small number of correct words (usually 0), or an
unacceptably large percentage of errors.

In this paper, we consider a completely new approach to obtaining a
``clean word list'' for document specific modeling. Rather than trying
to estimate the probability that an intermediate output of an OCR
system (like an HMM or CRF) is correct and then thresholding this
probability, we instead form a set of hypotheses about each word in
the document. Each hypothesis poses that one particular word of the
first-pass OCR system is incorrect. We then search for hypotheses that
we can reject with high confidence.  More formally, we treat a third
party OCR system (in this case, the open source OCR program Tesseract
\cite{tesseract:website}) as a null hypothesis generator, in which
each attempted transcription $T$ produced by the OCR system is treated
as the basis for a separate null hypothesis. The null hypothesis for
word $T$ is simply ``{\em Transcription $T$ is incorrect.}''.  Letting
$W$ be the true identity of a transcription $T$, we notate this as
$$
T \neq W.
$$

Our goal is to find as many hypotheses as possible that can be
rejected {\em with high confidence}. For the purposes of this
paper, we take high confidence to mean with fewer than 1 error in
a million rejected hypotheses. Since we will only be attempting to
reject a few hundred hypotheses, this implies we shouldn't make any
errors at all. As our results section shows, we do not make any errors
in our clean word lists, even when they come from quite challenging
documents. 

Before proceeding, we stress that the following are {\bf not} goals of this paper:
\begin{itemize}
\item to present a complete system for OCR,
\item to produce high area under a precision recall curve for all of the words in a document,
\item to produce accurate estimates of the probability of error of
particular words in OCR.
\end{itemize}
Once again, our goal is simply to produce large lists of clean words
from OCR output so that we can use them for document-specific
modeling.  After presenting our method for producing clean word lists,
we provide a formal analysis of the bounds on the probability of
incorrectly including a word in our clean word list, under certain
assumptions. When our assumptions hold, our error bound is very loose,
meaning our true probability of error is much lower. However, some
documents do in fact violate our assumptions. Finally, we give an
example of what can be done with the clean word lists by showing some
simple error correction results on a real-world OCR problem.
\section{Method for Producing Clean Word Lists}
In this section, we present our method for taking a document and the
output of an OCR system and producing a so-called {\em clean word
list,} i.e. a list of words which we believe to be correct, with
high confidence. Our success will be measured by the number
of words that can be produced, and whether we achieve a very low
error rate. A priori, we decided that our method would be a failure
if it produced more than a single error per document processed.
To be a success, we must produce a clean word list
which is long enough to produce an interesting amount of training
data for document-specific processing. 

We assume the following setup.
\begin{itemize}
\item We are provided with a document $D$ in the form of a grayscale image.
\item We are provided with an OCR program. From here forward, we will use
the open source OCR program Tesseract as our default OCR program.
\item We further assume that the OCR program can provide an {\em
  attempted} segmentation of the document $D$ into words, and that the
  words are segmented into characters. {\em It is not necessary that
    the segmentation be entirely correct, but merely that the program
    produces an attempted segmentation.}
\item In addition to a segmentation of words and letters, the program should
produce a best guess for every character it has segmented, and hence, by
extension, of every word (or string) it has segmented. Of course, we
do not expect all of the characters or words to be correct, as that
would make our exercise pointless.
\item Using the segmentations provided by the OCR program, we
assume we can extract the gray-valued bitmaps representing each guessed
character from the original document.
\item Finally, we assume we are given a lexicon. Our method is relatively robust
to the choice of lexicon, and assumes there will be a significant number of
non-dictionary words.
\end{itemize}

We define a few terms before proceeding. We define the {\em Hamming
  distance} between two strings of the same number of characters to be
the number of character substitutions necessary to convert one string
to the other. We define the {\em Hamming ball} of radius $r$ for a
word $W$, $H_r(W)$, to be the set of strings whose Hamming distance to
$W$ is less than or equal to $r$. Later, after defining certain
equivalence relationships among highly confusable characters such as
``o'' and ``c'', we define a {\em pseudo-Hamming distance} which is
equivalent to the Hamming distance except that it ignores
substitutions among characters in the same equivalence class. We also
use the notions of edit distance, which extends Hamming distance by
including joins and splits of characters, and pseudo-edit distance, which
is edit distance using the aforementioned equivalence classes.

Our method for identifying words in the clean list has three basic steps. We consider each word $T$ output by Tesseract.
\begin{enumerate}
\item First, if $T$ is not a dictionary word, we discard it and make
no attempt to classify whether it is correct. That is, we will not put
it on the clean word list.\footnote{Why is it not trivial to simply declare any output of an OCR program
that is a dictionary word to be highly confident? The reason is that
OCR programs frequently use language models to project uncertain
words onto nearby dictionary words. For example, suppose that 
rather than ``Rumplestiltskin'', the original string was ``Rumpledpigskin'',
and that the OCR program, confused by its initial interpretation, had projected
``Rumpledpigskin'' onto the nearest dictionary word ``Rumplestiltskin''. 
A declaration that this word was correct would then be wrong. However, 
our method will not fail in this way because if the true string were in
fact ``Rumpledpigskin'', the character consistency check would never
pass. It is for this reason that our method is highly non-trivial, and
represents a significant advance in the creation of highly accurate
clean word lists.
}
\item Second, given that $T$ is a dictionary word, we 
evaluate whether $H_1(T)$ is non-empty, i.e. whether there are 
any dictionary words for which a single change of a letter can produce
another dictionary word. If so, we discard the word, and again make
no attempt to classify whether it is correct.
\item Assuming we have passed the first two tests, we now perform a
  {\em consistency check} (described below) of each character in the
  word. If the consistency check is passed, we declare the word to be
  correct.
\end{enumerate}

\paragraph{Consistency check} 
In the following, we refer to the bitmap associated with a character
whose identity is unknown as a {\em glyph}. Let $W_j$ be the true
character class of the $j$th glyph of a word $W$, and let $T_j$ be the
Tesseract interpretation of the same glyph.  The goal of a consistency
check is to try to ensure that the Tesseract interpretation of a glyph
is reliable.  We will assess reliability by checking whether other
similar glyphs usually had the same interpretation by Tesseract.

To understand the purpose of the consistency check, consider the
following situation.  Imagine that a document had a stray mark that
did not look like any character at all, but was interpreted by
Tesseract as a character.  If Tesseract thought that the stray mark
was a character, it would have to assign it to a character class like
``t''.  We would like to detect that this character is unreliable. Our
scheme for doing this is to find other characters that are similar to
this glyph, and to check the identity assigned to those characters by
Tesseract. If a large majority of those characters are given the same
interpretation by Tesseract, then we consider the original character
to be reliable. Since it is unlikely that the characters closest to
the stray mark are clustered tightly around the true character ``t'',
we hope to detect that the stray mark is atypical, and hence
unreliable.

More formally, to test a glyph $g$ for reliability, we first 
find the $M$ characters in the document that are most similar to $g$.
We then run the following procedure:
\begin{itemize}
\item Step 1. Initialize $i$ to 1.  
\item Step 2. Record the class of the
character that is $i$th most similar to $g$. We use normalized correlation as
a similarity measure. 
\item Step 3. If any character
class $c$ has matched $g$ a number of times $c_n$ such that
$\frac{c_n}{i+1}>\theta$, then declare the character $g$ to be {\bf
  $\theta$-dominated} by the class $c$, and terminate the procedure.
\item Step 4. Otherwise, add 1 to i. If $i<M$, go to Step 2.
\item Step 5. If, after the top $M$ most similar characters to $g$
are evaluated, no character class $c$ dominates the glyph, then we
declare that the glyph $g$ is {\bf undominated}. 
\end{itemize}

There are three possible outcomes of the consistency check. The first
is that the glyph $g$ is dominated by the same class $c$ as the
Tesseract interpretation of $g$, namely $T_j$. The second outcome is
that $g$ is dominated by some other class that does not match
$T_j$. The third outcome is that $g$ is undominated. In the latter
two cases, we declare the glyph $g$ to be {\em unreliable}. Only if
the glyph $g$ is dominated by the same class as the Tesseract identity
of $g$ do we declare $g$ to be reliable. Furthermore, only if all of
the characters in a word are reliable do we declare the word to be
reliable or ``clean''.

The constants used in our experiments were $M=20$ and
$\theta=0.66$. That is, we compared each glyph against a maximum of 20
other glyphs in our reliability check, and we insisted that a
``smoothed'' estimate of the number of similarly interpreted glyphs
was at least 0.66\% before declaring a character to be reliable. We
now analyze the probability of making an error in the clean set.

\section{Theoretical bound}

For a word in a document, let $W$ be the ground truth label of the
word, $T$ be the Tesseract labeling of the word, and $C$ be a binary
indicator equal to 1 if the word passed the consistency check.  We
want to upper bound the probability $\Pr(W \neq w_t | T = w_t, C=1)$
when $w_t$ is a dictionary word and has an empty Hamming ball of size
1.

\begin{eqnarray}
\Pr(W \neq w_t | T = w_t, C=1) &=& \sum_{w \neq w_t} \Pr(W = w | T = w_t, C=1) \\
&=& \sum_{w \neq w_t, w \in \text{Dict}} \Pr(W = w | T = w_t, C=1) \\
&& + \sum_{w \neq w_t, w \notin \text{Dict}} \Pr(W = w | T = w_t, C=1) \notag \\
&=& \sum_{w \neq w_t, w \in \text{Dict}, |w| = |w_t|} \Pr(W = w | T = w_t, C=1) \\
&&  + \sum_{w \neq w_t, w \in \text{Dict}, |w| \neq |w_t|} \Pr(W = w | T = w_t, C=1) \notag \\
&&  + \sum_{w \neq w_t, w \notin \text{Dict}} \Pr(W = w | T = w_t, C=1) \notag
\end{eqnarray}

\subsection{Bounding the character consistency check}

We will rewrite the $\Pr(W = w | T = w_t, C = 1)$ terms as bounds
involving $\Pr(C = 1 | T = w_t, W = w)$ using Bayes' Rule.  We will
make the assumption that the individual character consistency checks
are independent, although this is not exactly true, since there may be
local noise that degrades characters in a word in the same way.  

Assume that each character is formed on the page by taking a single
true, latent appearance based on the font and the particular character
class and adding some amount of noise.  Let $\epsilon$ be an upper
bound on the probability that noise has caused a character of any
given class to look like it belongs to another specific class other
than its own class.  More formally, letting $p_c(a)$ be the
probability of a character appearance $a$ for a given class $c$ under
the noise model, $\epsilon$ satisfies, for all character classes $c_1,
c_2, c_1 \neq c_2$,

\begin{eqnarray}
\epsilon &>& \int_{a | p_{c_1}(a) < p_{c_2}(a)} p_{c_1}(a) da. \label{eq:eps-c}
\end{eqnarray}

In order to obtain a small value for $\epsilon$, and hence later a
small probability of error, we revise Eq.~\ref{eq:eps-c} to be a bound
only on \emph{non-confusable} character classes.  In other words,
since some character classes are highly confusable, such as 'o', 'c',
and 'e', we ignore such substitutions when computing Hamming and edit
distance.  We'll refer to these distances as modified distances, so
{\tt mode} and {\tt mere} have a true Hamming distance of 2 but a
modified Hamming distance of 1.

This is similar to defining an equivalence relation where confusable
characters belong to the same equivalence class, and computing
distance over the quotient set, but without transitivity, as, for
example, 'h' may be confusable with 'n', and 'n' may be confusable
with 'u', but 'h' may not necessarily be confusable with 'u'.

For a character to pass a consistency check with the label $c_2$ when
the true underlying label is $c_1$, roughly one of two things must
happen: (a) either the character was corrupted and looked more like
$c_2$ than $c_1$, or (b) some number of other characters with label
$c_2$ were corrupted and looked like $c_1$'s.

The probability (a) is clearly upper bounded by $\epsilon$, since it
requires both the corruption and most of its neighbors to have the
same label $c_2$.  Since $\epsilon \ll 1$ and (b) requires several
other characters with label $c_2$ to be corrupted to look like $c_1$,
the probability of (b) should be bounded by (a), and thus $\epsilon$,
as well.

Therefore the probability of the consistency check giving a label
$c_2$ when the true underlying label is $c_1$ is less than
$2\epsilon$, for any classes $c_1, c_2$.

We will also need a lower bound on the probability that a character
consistency check will succeed if the Tesseract label of the character
matches the ground truth label.  Let $\delta$ be a lower bound on this
quantity, which is dependent on both the amount of noise in the
document and the length of the document.  (The latter condition is due
to the fact that the character consistency check requires a character
to match to at least a certain number of other similarly labeled
characters, so, for example, if that number isn't present in the
document to begin with, then the check will fail with certainty.)

\subsection{Bounding one term}

Consider, bounding $\Pr(W = w | T = w_t, C = 1)$:

\begin{eqnarray}
&& \Pr(W = w | T = w_t, C = 1) \notag \\ 
&& \qquad \qquad = \frac{\Pr(C = 1|T=w_t, W=w) \Pr(W = w | T = w_t)}{\sum_{w'} \Pr(C = 1|T=w_t, W=w') \Pr(W = w' | T = w_t)} \\
&& \qquad \qquad = \frac{\Pr(C = 1|T=w_t, W=w) \Pr(T = w_t | W = w) \Pr(W = w)}{\sum_{w'} \Pr(C = 1|T=w_t, W=w') \Pr(T = w_t | W = w') \Pr(W = w')} \\
&& \qquad \qquad \leq \frac{\Pr(C = 1|T=w_t, W=w) \Pr(T = w_t | W = w) \Pr(W = w)}{\Pr(C = 1|T=w_t, W=w_t) \Pr(T = w_t | W = w_t) \Pr(W = w_t)} \label{eq:a}
\end{eqnarray}

\subsection{Bounding dictionary words}

For dictionary words $w$, we will assume that

\begin{eqnarray}
\frac{\Pr(T = w_t | W = w) \Pr(W = w)}{\Pr(T = w_t | W = w_t) \Pr(W = w_t)} &<& 1,\label{eq:balance}
\end{eqnarray}

\noindent or, equivalently,

\begin{eqnarray}
\frac{\Pr(W = w | T = w_t)}{\Pr(W = w_t | T = w_t)} &<& 1.
\end{eqnarray}

Eq.~\ref{eq:balance} can be thought of as balance between Tesseract's
word level accuracy $\frac{\Pr(T = w_t | W = w)}{\Pr(T = w_t | W =
  w_t)}$ and the words' prior probabilities $\frac{\Pr(W = w)}{\Pr(W =
  w_t)}$.  In our documents, word level accuracy $\Pr(T = w_t | W =
w_t)$ is generally around .9, and the numerator in the first ratio
$\Pr(T = w_t | W = w)$ is the probability of Tesseract making a
specific mislabeling, and so is upper bounded by .1 but generally will
be much lower.

Applying this to Eq.~\ref{eq:a}, we get

\begin{eqnarray}
\Pr(W = w | T = w_t, C = 1)
&\leq& \frac{\Pr(C = 1|T=w_t, W=w)}{\Pr(C = 1|T=w_t, W=w_t)}. \label{eq:b}
\end{eqnarray}

\subsubsection{Bounding dictionary Hamming words}

Consider a word $w$ that is a modified Hamming distance $i$ from
$w_t$.  We can then simplify Eq.~\ref{eq:b} as

\begin{eqnarray}
\Pr(W = w | T = w_t, C = 1)
&\leq& \frac{(2\epsilon)^i}{\delta^i}
\end{eqnarray}

\noindent by making use of the assumption that the character
consistency checks are independent, and that $w$ and $w_t$ only differ
in $i$ characters.  For those $i$ characters, $w$ does not match the
Tesseract label and $w_t$ does match the Tesseract label, so we use
the bounds $2\epsilon$ and $\delta$.

Now let $D_i$ be the number of dictionary words of modified Hamming
distance $i$ away from $w_t$.  Let $r_D$ be the rate of growth of
$D_i$ as a function of $i$, e.g. $D_{i+2} = r_D^i D_2$.  Assume, since
$\epsilon \ll 1$, that $r_D(\frac{2\epsilon}{\delta}) <
\frac{1}{2}$.  (Experiments on the dictionary used in our main
experiments showed that $r_D$ is generally about 3.)

To get the total contribution to the error from all dictionary Hamming
words, we sum over $D_i$ for all $i > 1$,

\begin{eqnarray}
\sum_{w \neq w_t, w \in \text{Dict}, |w| = |w_t|} \Pr(W = w | T = w_t, C=1) 
&\leq& \sum_{i=2} D_i \frac{(2\epsilon)^i}{\delta^i} \\
&=& D_2 \frac{(2\epsilon)^2}{\delta^2} + D_2 \frac{(2\epsilon)^2}{\delta^2} \sum_{i=1} (2r_D\frac{\epsilon}{\delta})^i \\
&\leq& 8D_2 \frac{\epsilon^2}{\delta^2}.
\end{eqnarray}

\subsubsection{Bounding dictionary edit words}

Traditionally, edit distance is computed in terms of number of
substitutions, insertions, and deletions necessary to convert one
string to another string.  In our context, a more natural notion may
be splits and joins rather than insertions and deletions.  For
example, the interpretation of an 'm' may be split into an 'r' and a
'n', or vice-versa for a join.

The probability that a split or a join passes the consistency check is
upper bounded by $(2\epsilon)^2$.  We can see this from two
perspectives.  First, a split or join has traditional edit distance of
2, since it requires an insertion or deletion and a substitution ('m'
to 'mn' insertion followed by 'mn' to 'rn' substitution).  

A more intuitive explanation is that, for a split, one character must
be corrupted to look like the left hand side of the resulting
character and another character corrupted to look like the right hand
side, and for a join, the left hand side of a character must be
corrupted to look like one character and the right hand side corrupted
to look like another.

Similar to the case of confusable characters for substitutions, we
also ignore confusable characters for splits and joins, namely 'r'
followed by 'n' with 'm', and 'v' followed by 'v' with 'w'.  Thus,
{\tt corn} and {\tt comb} have an edit distance of 2 but a modified
edit distance of 1.

Consider a word $w$ with modified edit distance $i$ from $w_t$, and
involving at least one insertion or deletion (so $|w| \neq |w_t|$).
Similar to the dictionary Hamming words, we can simplify
Eq.~\ref{eq:b} for $w$ as

\begin{eqnarray}
\Pr(W = w | T = w_t, C = 1)
&\leq& \frac{(2\epsilon)^{i+1}}{\delta^i},
\end{eqnarray}

\noindent since each substitution contributes a
$\frac{2\epsilon}{\delta}$ and each insertion or deletion, of which
there is at least one, contributes a $\frac{(2\epsilon)^2}{\delta}$.

Let $E_i$ be the number of dictionary words $w$ with a modified edit
distance $i$ away from $w_t$ and $|w| \neq |w_t|$.

Again, also assume that $r_E$, the rate of growth of $E_i$, satisfies
$r_E(\frac{2\epsilon}{\delta}) < \frac{1}{2}$.  Summing the total
contribution to the error from dictionary edit words,

\begin{eqnarray}
\sum_{w \neq w_t, w \in \text{Dict}, |w| \neq |w_t|} \Pr(W = w | T = w_t, C=1) 
&\leq& \sum_{i=1} E_i \frac{(2\epsilon)^{i+1}}{\delta^i} \\
&=& E_1 \frac{(2\epsilon)^2}{\delta} + E_1 \frac{(2\epsilon)^2}{\delta} \sum_{i=1} (2r_E\frac{\epsilon}{\delta})^i \\
&\leq& 8E_1 \frac{\epsilon^2}{\delta} \\
&\leq& 8E_1 \frac{\epsilon^2}{\delta^2}.
\end{eqnarray}

\subsection{Bounding non-dictionary words}

Let $N_i$ be the set of non-dictionary words with a modified edit
distance $i$ from $w_t$, and let $p_i = \frac{\Pr(T = w_t | W \in N_i)
  \Pr(W \in N_i)}{\Pr(T = w_t | W = w_t) \Pr(W = w_t)}$.  Assume the
rate of growth of $r_N$ of $p_i$ satisfies $r_N
(\frac{2\epsilon}{\delta}) < \frac{1}{2}$.

Rearranging Eq.~\ref{eq:a} and summing over all non-dictionary words:

\begin{eqnarray}
&& \sum_{w \neq w_t, w \notin \text{Dict}} \Pr(W = w | T = w_t, C=1) \notag \\
&& \qquad \qquad \leq \sum_{i=1} \sum_{w \in N_i} \frac{\Pr(C = 1|T=w_t, W=w) \Pr(W=w | T=w_t)}{\Pr(C = 1|T=w_t, W=w_t) \Pr(W=w_t | T=w_t)} \\
&& \qquad \qquad \leq \sum_{i=1} \sum_{w \in N_i} \frac{(2\epsilon)^i}{\delta^i} \frac{\Pr(W = w | T=w_t)}{\Pr(W=w_t | T=w_t)} \\
&& \qquad \qquad = \sum_{i=1} \frac{(2\epsilon)^i}{\delta^i} \frac{\Pr(W \in N_i | T=w_t)}{\Pr(W=w_t | T=w_t)} \\
&& \qquad \qquad = \sum_{i=1} \frac{(2\epsilon)^i}{\delta^i} p_i \\
&& \qquad \qquad \leq p_1 \frac{2\epsilon}{\delta} + p_1 \frac{2\epsilon}{\delta} \sum_{i=1} (2r_N \frac{\epsilon}{\delta})^i \\
&& \qquad \qquad \leq 4 p_1 \frac{\epsilon}{\delta^2}
\end{eqnarray}

\subsection{Final bound}

\begin{eqnarray}
\Pr(W \neq w_t | T = w_t, C = 1) &\leq& \frac{(8D_2 + 8E_1)\epsilon^2 + 4p_1 \epsilon}{\delta^2}
\end{eqnarray}

For $\epsilon < 10^{-3}$, $8D_2 + 8E_1 < 10^2$, $4p_1 < 10^{-1}$, $\delta^2 > 10^{-1}$,

\begin{eqnarray}
\Pr(W \neq w_t | T = w_t, C = 1) &\leq& 2\cdot 10^{-3}.
\end{eqnarray}

The bounds for the constants chosen above were selected conservatively
to hold for a large range of documents, from very clean to moderately
noisy.  Not all documents will necessarily satisfy these bounds.  In a
sense, these inequalities define the set of documents for which our
algorithm is expected to work, and for heavily degraded documents that
fall outside this set, the character consistency checks may no longer
be robust enough to guarantee a very low probability of error.

Our final bound on the probability of error, 0.002, is the
result of a {\em worst case analysis} under our assumptions. If our
assumptions hold, the probability of error will likely be much lower
for the following reasons.  For most pairs of
letters, $\epsilon = 10^{-3}$ is not a tight upper bound.  The
quantity on the right of Eq.~\ref{eq:balance} is typically much lower
than 1.  The rate of growths $r_D, r_E, r_N$ are typically much lower
than assumed.  The bound on $p_1$, the non-dictionary word
probabilities, is not a tight upper bound, as non-dictionary words
mislabeled as dictionary words are rare.  Finally, the number of
Hamming and edit distance neighbors $D_2$ and $E_1$ will typically be
less than assumed.

On the other hand, for sufficiently noisy documents, and certain types
of errors, our assumptions do not hold. Some of the problematic cases
include the following.  As discussed, the assumption that the
individual character consistency checks are independent is not true.
If a document is degraded or has a font such that one letter is
consistently interpreted as another, then that error will likely pass
the consistency check (e.g. $\epsilon$ will be very large).  If a
document is degraded or is very short, then $\delta$ may be much
smaller than $10^{-\frac{1}{2}}$.  (The character consistency check
requires a character to match to at least a certain number of other
similarly labeled characters, so, for example, if that number isn't
present in the document to begin with, then the check will fail with
certainty.)  Finally, if the dictionary is not appropriate for the
document then $4p_1 < 10^{-1}$ may not hold.  This problem is
compounded if the OCR system projects to dictionary words.

\section{Experiments}
\begin{figure*}[t!]
\centering
\includegraphics[width=1.\linewidth]{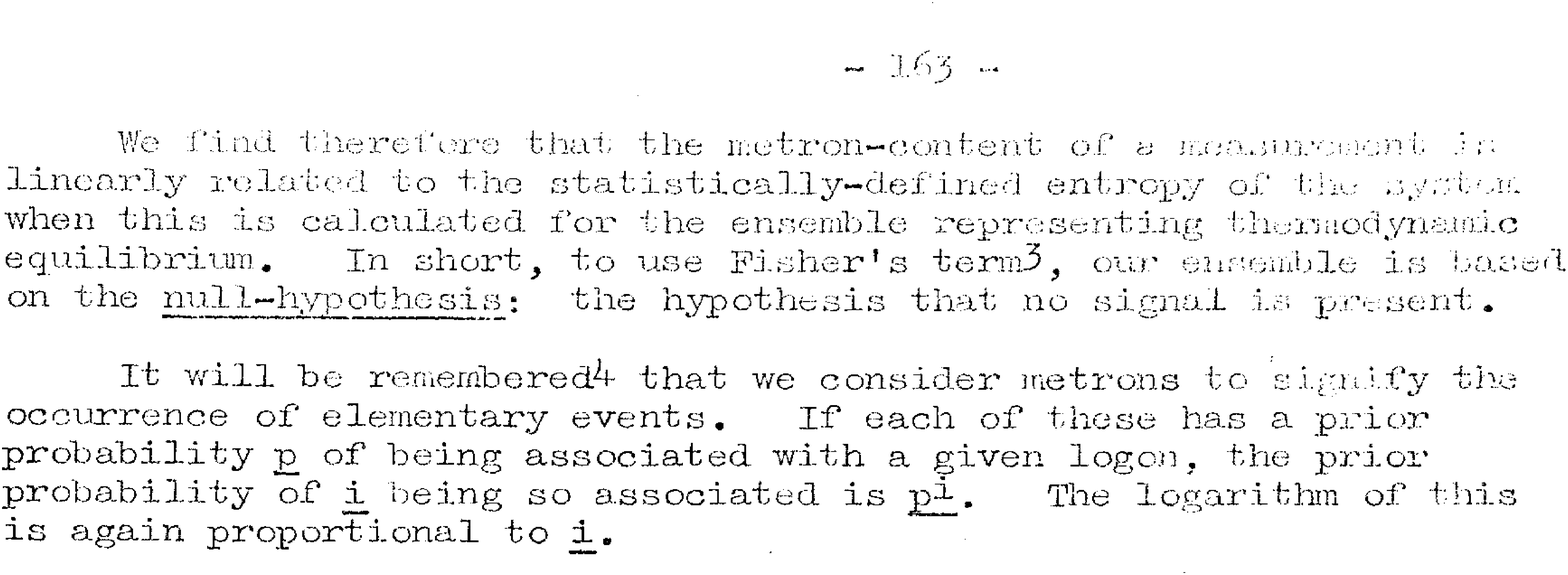}
\caption{The top portion of page 2 of the Mackay document. Notice the degraded characters, which produces a significant number of errors in the Tesseract OCR program.}
\label{fig:Mackay}
\end{figure*}

We analyzed 10 short documents (4-5 pages each) with Tesseract word
accuracy rates between 58\% and 95\% (see Table 1) \cite{mackay53,archie:gutenberg, oahu:gutenberg, history53, campfire:gutenberg, beauty:gutenberg, cristo:gutenberg, country:gutenberg, geology:UNLV_corpus, cosmic:gutenberg}. A portion of one
document is shown in Figure~\ref{fig:Mackay}.  Note that in order to simulate document degradation, we downsampled some of the documents from the Project Gutenberg corpus.  

Applying the method of
Section~3, we produced ``conservative'' clean word lists as shown in
Table 1. {\bf Our main result is that we are
  able to extract an average of 6\% of the words in each document
  without making a single error.}
The total number of words in all of the conservative clean lists was
843, giving some informal support to the legitimacy of our bound,
which suggests we would expect between 1 and 2 errors in a worst case
scenario, and fewer than that under average conditions.  We found that
we could relax the requirement that a word have empty Hamming ball of
radius 1 (Step 2 of our method) and replace it with a less stringent
requirement, while only introducing a single error into what we called
an ``aggressive clean word list'' (see Table 1). In particular, we
replaced the empty Hamming ball requirement with a requirement that a
word either have an empty Hamming ball of radius 1, {\em or}, if it
had neighbors of Hamming distance 1, that those neighbors not occur
elsewhere in the document. This less restrictive definition of the
clean word sets, no longer subject to our analytic bound,
produces much larger clean lists (2661 words) while
producing just a single error.

\begin{table*}
\begin{centering}\begin{tabular}{|c|c|c|c|c|c|c|c|c|}
  \hline
  Document &
  Word &
  Tesseract&
  \multicolumn{2}{|c|}{Conservative List} &
  Conserv. &
  \multicolumn{2}{|c|}{Aggressive List} &
  Aggress. \\
Name &
Count &
Accuracy &
Count &
\% &
Errors &
Count &
\% &
Errors\\
  \hline
  Archie &
  754 &
  80.1\% &
  29  & 3.8\% &
  \bf{0} &
  106 & 14.1\% &
  \bf{0} \\
  \hline
  Beauty &
  1706 &
  86.1\% &
  69  & 4.0\% &
  \bf{0} &
  308 & 18.1\% &
  \bf{0} \\
  \hline
  Campfire &
  711 &
  64.3\% &
  4 & 0.6\% &
  \bf{0} &
  27 & 3.8\% &
  \bf{0} \\
  \hline
  Cosmic & 1533 &    86.6\% &    60 & 3.9\% &    \bf{0} &   222 & 14.5\% &    \bf{1} \\
  \hline
  Country &
  892 &
  58.3\% &
  20 & 2.2\%  &
  \bf{0} &
  92 & 10.3\% &
  \bf{0} \\
  \hline
  Cristo &
  886 &
  86.7\% &
  31  & 3.5\% &
  \bf{0} &
  136 & 15.3\% &
  \bf{0} \\
  \hline
  Geology &
  1498 &
  95.2\% &
  121 & 8.1\%  &
  \bf{0} &
  392 & 26.2\% &
  \bf{0} \\
  \hline
  History &
  2712 &
  79.7\% &
  245 & 9.0\%  &
  \bf{0} &
  571 & 21.1\% &
  \bf{0} \\
  \hline
  Mackay &
  2056 &
  85.4\% &
  175  & 8.5\% &
  \bf{0} &
  522 & 25.4\% &
  \bf{0} \\
  \hline
  Oahu &
  1309 &
  88.9\% &
  89 & 6.8\%   &
  \bf{0} &
  285 & 21.8\% &
  \bf{0} \\
  \hline \hline
Totals & 14057 & 82.9\% & 843 & 6.0\% & \bf{0} &  2661 & 18.9\% & \bf{1} \\ \hline
\end{tabular}\par\end{centering}
\caption{Results. Using our conservative clean list approach, we
  retrieved an average of 6\% of the words in our test documents with
  0 errors. Using our aggressive approach, we obtained an average of
  18.9\% with only a single error.\label{tab:accuracy}}
\end{table*}

As mentioned in the introduction, many OCR programs come with some
notion of a confidence estimate for each word or character that is
produced. Tesseract has two separate functions which are related to
its confidence of word accuracy. An obvious approach to producing
clean word lists is to simply learn a threshold on the confidence
values output by an OCR program.

\begin{figure}[t]
\includegraphics[width=\textwidth]{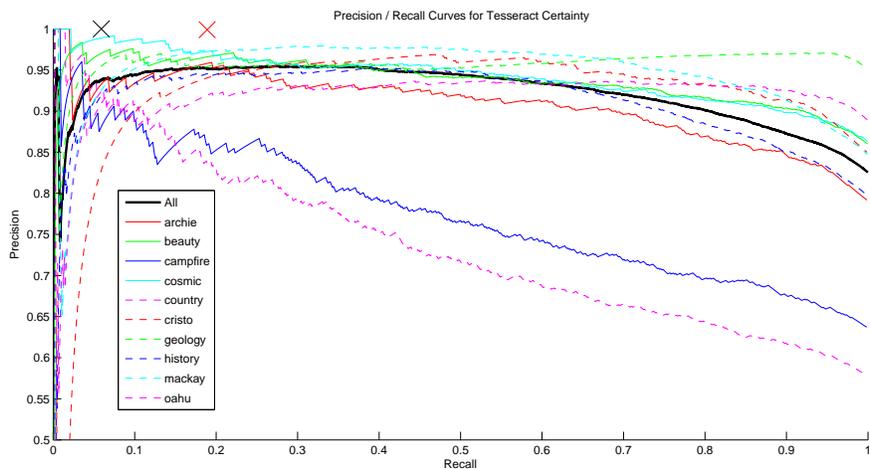}
\caption{Precision recall curves using Tesseract's confidence measure for individual documents (colored lines) and aggregate document (heavy black line). No threshold on the confidence measure produces satisfactory clean word lists. Our points on the precision recall curve are shown for our conservative procedure (black X on left) and our aggressive procedure (red X on right).}
\label{fig:means}
\end{figure}

Consider the heavy black line in Figure~\ref{fig:means}. This line
shows the precision/recall curve for the words in {\em all} of our
test documents together as the threshold on the Tesseract confidence
value is changed. It is easy to see that there is no threshold of the
confidence value that achives a word accuracy rate of higher than
about 95\%. Hence, there is no confidence level which can be used to
produce clean word lists across all documents. The individual colored
curves show precision/recall curves for individual documents. While
there exist Tesseract confidence thresholds that produce error-free
clean word lists for some documents, the same confidence levels do not
work for other documents, showing, that as a general technique, using
a threshold on the Tesseract confidence value cannot be used in a
naive fashion to produce such clean word lists.

Finally, notice the black and red X'es at the top of the figure. The
black X (on the left) represents the average recall rate of about 6\% and the
perfect precision of our conservative clean word generation scheme
averaged across all documents. The red X (on the right) shows the slightly lower
precision and significantly higher recall of our aggressive clean word
generation scheme, again averaged across all documents.

\subsection{Using the clean word sets}
While it is not the main thrust of this paper, we have begun to
experiment with using the clean word sets to improve OCR results. In
our first experiment, we extract all examples of a pair of confusable
characters (``o'' and ``c'') from the conservative list of the Mackay document. We then trained
a linear support vector machine classifier on this list,
and reclassified all instances of ``o'' and ``c'' not in the conservative
list. While Tesseract made four errors
confusing ``o''s and ``c''s in the original document, all of these
were corrected using our document-specific training set, and no new
errors were introduced. While this clearly represents just a
beginning, we believe it is a promising sign that document specific
training sets will be useful in correcting a variety of errors in OCR
documents. It is interesting to note that no new data needed to be labeled
for this experiment. Everything was done fully automatically.

\paragraph{Related work.}
There has been significant work done in post-processing the output of
OCR.  While none of it is very similar to what we have presented here,
we present some points of reference for completeness.  Kolak et
al.~\cite{Kolak03agenerative} developed a generative model to estimate
the true word sequence from noisy OCR output and Kukich et
al.~\cite{146380} survey various methods to correct text using
natural language processing techniques.  Hong et
al.~\cite{Hong95visualinter-word} examine the inter-word
relationships of character patches to help constrain possible
interpretations.  The distinguishing feature of our work is that we
examine the document images themselves to build document-specific
models of the characters.

Our work is also related to a variety of approaches that
leverage inter-character similarity in documents in order to reduce
the dependence upon a priori character models. Some of the most
significant developments in this area are found in these papers
\cite{Casey1986,Breuel2003,Lee2002,Nagy1986,HobbyHo1997,HoNagy2000}.
The inability to attain high confidence in either the identity or 
equivalence of characters in these papers has hindered their use
in subsequent OCR developments. We hope that the high confidence
values we obtained will spur the use of these techniques for document-specific
modeling.

\newpage
\small{
\bibliographystyle{plain}
\bibliography{/home/drew/workspace/trunk/vision/OCR/papers/bibtex/akae.bib,/home/drew/workspace/trunk/vision/OCR/papers/bibtex/emiller02.bib}
}

\end{document}